\documentclass[conference]{IEEEtran}
\IEEEoverridecommandlockouts
\usepackage{cite}
\usepackage{amsmath,amssymb,amsfonts}
\usepackage{algorithmic}
\usepackage{graphicx}
\usepackage{textcomp}
\usepackage{xcolor}
\def\BibTeX{{\rm B\kern-.05em{\sc i\kern-.025em b}\kern-.08em
    T\kern-.1667em\lower.7ex\hbox{E}\kern-.125emX}}
\begin{document}

\makeatletter
\newcommand{\newlineauthors}{%
  \end{@IEEEauthorhalign}\hfill\mbox{}\par
  \mbox{}\hfill\begin{@IEEEauthorhalign}
}
\makeatother

\title{A Novel end-to-end Framework for Occluded Pixel Reconstruction with Spatio-temporal Features for Improved Person Re-identification\\
}



\author{

\IEEEauthorblockN{Prathistith Raj Medi}
\IEEEauthorblockA{\textit{IIIT Naya Raipur}\\
prathistith19102@iiitnr.edu.in}
\and

\IEEEauthorblockN{Ghanta Sai Krishna}
\IEEEauthorblockA{\textit{IIIT Naya Raipur}\\
ghanta20102@iiitnr.edu.in}
\and

\IEEEauthorblockN{Praneeth Nemani}
\IEEEauthorblockA{\textit{IIIT Naya Raipur}\\
praneeth19100@iiitnr.edu.in}

\newlineauthors
\IEEEauthorblockN{Satyanarayana Vollala}
\IEEEauthorblockA{\textit{IIIT Naya Raipur}\\
satya@iiitnr.edu.in}
\and
\IEEEauthorblockN{Santosh Kumar}
\IEEEauthorblockA{\textit{IIIT Naya Raipur}\\
santosh@iiitnr.edu.in}
\textit{ORCID: 0000-0003-2264-9014}

}

\IEEEoverridecommandlockouts
\IEEEpubid{\makebox[\columnwidth]{978-1-5386-5541-2/18/\$31.00~\copyright2018 IEEE \hfill}
\hspace{\columnsep}\makebox[\columnwidth]{ }}
\maketitle
\IEEEpubidadjcol

\begin{abstract}
Person re-identification is vital for monitoring and tracking crowd movement to enhance public security. However, re-identification in the presence of occlusion substantially reduces the performance of existing systems and is a challenging area. In this work, we propose a plausible solution to this problem by developing effective occlusion detection and reconstruction framework for RGB images/videos consisting of Deep Neural Networks. Specifically, a CNN-based occlusion detection model classifies individual input frames, followed by a Conv-LSTM and Autoencoder to reconstruct the occluded pixels corresponding to the occluded frames for sequential (video) and non-sequential (image) data, respectively. The quality of the reconstructed RGB frames is further refined and fine-tuned using a Conditional Generative Adversarial Network (cGAN). Our method is evaluated on four well-known public data sets of the domain, and the qualitative reconstruction results are indeed appealing. Quantitative evaluation in terms of re-identification accuracy of the Siamese network showed an exceptional Rank-1 accuracy after occluded pixel reconstruction on various datasets. A comparative analysis with state-of-the-art approaches also demonstrates the robustness of our work for use in real-life surveillance systems.
\end{abstract}


\section{Introduction}
Person re-identification refers to finding one-to-one correspondence between person images captured by a pair of cameras with non-overlapping fields of view \cite{24}. 
A network of surveillance cameras are installed for monitoring in most public spaces, such as railway stations, airports, shopping malls, hospitals, and office buildings. Manual analysis of this large amount of video data to perform person re-identification or other video surveillance tasks is laborious and time-intensive.
In this work, we propose an automated computer vision-based approach for re-identification for both image and video data in the presence of occlusion. 
Several re-identification approaches have been developed that operate on unoccluded frames \cite{7}. However, in most real-life situations, occlusion is an inevitable occurrence. It emerges when static or dynamic objects appear between the camera field-of-view and the target subject, causing certain regions of the target subject to get obstructed. 
The presence of occlusion degrades the performance of traditional image-based re-identification techniques, and the problem is likely to amplify if multiple consecutive frames in a video are occluded. Previous approaches to handling occlusion in re-identification attempt to leverage pose estimation and visible areas i.e., the spatial information of unoccluded pixels in the frame while ignoring the temporal relation between adjacent frames \cite{3, 4}. To date, not much focus has been given to occluded pixel reconstruction in video-based person re-identification with temporal features.
In this work, we address a plausible solution to this problem by proposing a novel multi-model architecture for occlusion reconstruction from both image and video data. For video or sequential data, we propose a \textit{Conv-LSTM} model to reconstruct the occluded pixels in a frame by exploiting the spatio-temporal information from previous frames followed by fine-tuning with \emph{cGAN} \cite{6}. For image data, we propose an \emph{Autoencoder} model for occlusion reconstruction followed by the same \emph{cGAN}-based fine-tuning to enhance the reconstruction further. Employing the \emph{cGAN}-based fine-tuning stage helps reduce noise and artifacts, thereby preserving better translation in-variance in the reconstructed frames or images generated by the previous networks. 
Finally, a siamese network-based re-identification framework has been used to perform the identity matching. The overall multi-model occluded pixel reconstruction re-identification strategy has been detailed in Section 3. 

\begin{figure*}[t]
    \centering
    \includegraphics[width=\textwidth]{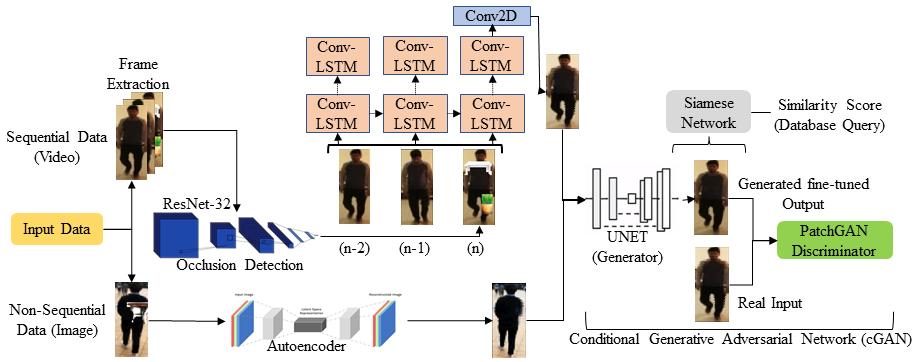} 
    \caption{Overview of the proposed Occluded pixel Reconstruction and Person Re-identification Framework for both Image and Video data with Spatial and Spatio-temporal modelling respectively}
    \label{overview}
\end{figure*}

\noindent
To summarize the main contributions of the paper are as follows:
\begin{enumerate}
    \item Occlusion detection \& reconstruction for efficient re-identification using advanced deep learning techniques is a significant contribution of this work. To the best of our knowledge, re-identification from video data consisting of sequential frames with occlusion reconstruction using spatio-temporal features has not been emphasized much in the past, which we aspired to solve here.
    
    \item Practical approaches to occlusion reconstruction using a multi-modal framework have been proposed for both sequential (video) and non-sequential (image) data. A \textit{Conv-LSTM} and \emph{Autoencoder} for video and image data, respectively, followed by \emph{cGAN} to generate a visually appealing reconstruction in occluded regions, is proposed. 
    \item An extensive experimental evaluation and comparative analysis of our method with other state-of-the-art approaches using four publicly available data sets is accomplished. 

\end{enumerate}


\section{Background and Related Work}\label{sec 2}
This section presents an overview of the existing approaches to person re-identification broadly: still images, sequential frames, i.e., videos, and occlusion.

The methods including re-identification from still images rely entirely on visual descriptors and do not incorporate any external contextual information for establishing correspondence between images \cite{7}.
In contrast recently, several deep learning-based person re-identification techniques have been developed by researchers worldwide \cite{8, 9}. 
The approach proposed in \cite{8} presents a two-stream network, namely \textit{OSNet}, to address discriminative person-specific learning with generalizable property for cross-dataset discrepancies by employing  instance normalization layers in \textit{OSNet}. The work proposed in \cite{9}, solved the wrong label and poor quantity data problem where a weighted label correction based on cross-entropy is introduced to solve wrong labeling and to correct the two label errors weighted triplet loss. 

Previous works on temporal modeling methods for video-based person re-identification use sequential models like recurrent neural networks (RNN).
In \cite{10}, McLaughlin et al. first introduced the concept of modeling temporal information from frames using a RNN in which the average of RNN cell outputs is used as clip level representations. 
Further, an unsupervised approach for improved label estimation in person re-identification is presented based on a dynamic graph matching (DGM) framework in which intermediate labels are used to iteratively refine the framework for better labeling \cite{12}.
A modality-aware collaborative ensemble learning method is employed  in \cite{14}  to handle the modality discrepancy at both the feature-level and the classifier-level during re-identification. Attention-based models have also been used in person re-identification to extract spatio-temporal features from video sequences A multi-attention convolutional neural network (MA-CNN) for part localization and fine-grained feature learning is presented by Zheng et al. in \cite{15}. While \textit{Co-attention Siamese Network} (COSNet)   segments and encodes useful image features, 
the work in \cite{16}  addresses the challenging problem of saliency shift by employing saliency-shift-aware \textit{Conv-LSTM} layer which can efficiently capture video saliency dynamics through learning human attention-shift behavior. However, none of these approaches are specifically tuned to perform re-identification in the presence of occlusion.

Gao et al. \cite{17} proposed a pose-guided Visible Part Matching (PVPM) model to learn attentions with discriminative part features. In  a cross-graph embedded-alignment (CGEA) layer is used to embed topology information to local features and predict similarity scores for matching. A solution to tackle the degradation of recognition performance caused by occlusion is addressed in  in which the authors proposed a two-step solution: (i) extract non-occluded human body features through pose estimation, and (ii) locate person body parts by utilizing the detected human keypoints in different occlusion situations.  
 
From the extensive literature survey, we observe that person re-identification by exploiting the spatio-temporal information from sequential video data for occlusion reconstruction and re-identification has not been addressed in the literature. Also, the quality of  reconstructed frames by the existing models using only spatial features needs to be improved significantly. In this work, we propose novel and effective approaches to occlusion detection and reconstruction for both sequential and non-sequential image frames with a focus on enhancing the quality of the reconstruction thereby achieve a better re-identification performance.

\section{Proposed Work}\label{sec 3}
The proposed work has been presented in two modules detailing the occlusion detection and the proposed Multi-modal framework for occlusion reconstruction and  re-identification with Siamese network. These modules are explained in the following three sub-sections.

\subsection{Multi-modal Framework for Occlusion Reconstruction}
Initially for occluded frame detection we employ a CNN with ResNet-34 architecture to classify if a given frame is occluded or not. We essentially use it as a binary classifier with the last layer consisting sigmoid activation function with the outcome '1' or '0' representing `occluded' and `un-occluded' classes respectively while the input to the classifier is an RGB frame. 
The ResNet-34 is trained using synthetically occluded images and un-occluded images with the corresponding labels. Adam optimizer is used to train the model using binary cross-entropy loss (see Eq \ref{eq1}) considering a learning rate of \textit{0.001}.
\begin{equation}
\small
    L_{bce}=-\frac{1}{N}\sum_{i=1}^{N}Y_{i} log(p(Y_{i}))+(1-Y_{i}) log(1-p(Y_{i}))
\label{eq1}    
\end{equation}
In (\ref{eq1}), $Y_{i}$ and $(1-Y_{i})$ represent two possible classes with target probabilities and N stands for the number of training samples. 

Further, occlusion reconstruction is carried out for only those frames characterized as `occluded' by the Occlusion Detection model. To reconstruct an occluded frame in a video sequence, we follow a two-step process: (i) first
a \textit{Conv-LSTM} is employed to reconstruct the occluded frame or region using the spatio-temporal features from previous frames, and (ii) next the predicted frame is fine-tuned using a Conditional Generative Adversarial Network (\textit{cGAN}) \cite{6}. The use of \textit{Conv-LSTM} in the first step helps in fair reconstruction of the occluded pixels with the spatio-temporal information. However, the image generated by \textit{Conv-LSTM} contains minor irregularities and noise. To refine further in producing visually appealing reconstructed frames, \textit{cGAN} has been employed on top of \textit{Conv-LSTM} as it is known to have  demonstrated robustness in handling image translation tasks.
In case non-sequential frames are available instead of a video sequence, \textit{Conv-LSTM} has not been used for occlusion reconstruction since the spatio-temporal information cannot be exploited in such cases. Instead, occlusion reconstruction in each occluded frame is accomplished using an Autoencoder network. The reconstructed frame is henceforth used to perform re-identification using a Siamese network. We next describe in detail each of the networks.

\subsubsection{Coarse Occlusion Reconstruction Using \textit{Conv-LSTM} / \textit{Autoencoder}}\label{convlstm}

\emph{Conv-LSTM} is a specially modified version of \textit{LSTM} (Long Short-Term Memory) \cite{18} in which the convolution operation replaces the matrix multiplication inside the \textit{LSTM} cell at every gate. This network is capable of capturing spatial features and learn long-term dependencies over time from sequential multi-dimensional data \cite{19}. The following equations show the various operations involved in the \emph{Conv-LSTM} layer. Here, $i_{t}$, $f_{t}$ and $o_{t}$ represent input, forget, and output gates respectively, and $X_{1} \dots X_{t}$ represent inputs to the layer while $C_{1} \dots C_{t}$ and $H_{1} \dots H_{t}$ represent cell outputs and hidden states, respectively, and $W$s are the weight matrices. The symbols $\ast$ and $\circ$ represent the convolution operator and Hadamard Product, respectively. 
\begin{equation}
\small
    i_{t}= \sigma (W_{xi}\ast X_{t}+W_{hi}\ast H_{t-1}+W_{ci}\circ  C_{t-1}+b_{i})
\end{equation}
\begin{equation}
\small
    f_{t}= \sigma (W_{xf}\ast X_{t}+W_{hf}\ast H_{t-1}+W_{cf}\circ  C_{t-1}+b_{f})
\end{equation}
\begin{equation}
\small
    C_{t}= f_{t} \circ C_{t-1} + i_{t} \circ tanh(W_{xc}\ast X_{t}+W_{hc}\ast H_{t-1}+b_{c}) 
\end{equation}
\begin{equation}
\small
    o_{t}= \sigma (W_{xo}\ast X_{t}+W_{ho}\ast H_{t-1}+W_{co}\circ  C_{t}+b_{o})
\end{equation}
\begin{equation}
\small
    H_{t}=o_{t} \circ tanh (C_{t})
\end{equation}

An insight view of the \textit{Conv-LSTM} architecture is given in Fig \ref{overview} and Table \ref{Table:convlstm} presents the layer-wise architecture of the model. Also, an occluded frame (say, \textit{Frame n}) is predicted by fusing the spatio-temporal information given by the respective occluded frame (i.e., \textit{Frame n}) along with the previous frames (i.e., \textit{Frame n-1} and \textit{Frame n-2}).

\begin{table}[ht]
	\scriptsize
	\centering
	\caption{Layer-wise specification of the \textit{Conv-LSTM}. ConvLSTM2d\_i represents the $i^{th}$ layer of the model} 
	\begin{tabular}{|c|c|c|c|}
		\hline
		\textbf{Network} & \textbf{Layer} & \textbf{Filter size} & \textbf{No. of filters} \\ \hline
		 & ConvLSTM2d\_1  & 5$\times$5 & 128 \\ \cline{2-4}
		 & ConvLSTM2d\_2 to & 3$\times$3 & 128,64,64, \\ 
		\emph{Conv-LSTM} & ConvLSTM2d\_8 &  & 32,32,32,16 \\
		\cline{2-4}
		 \emph{Model}& ConvLSTM2d\_9 & 1$\times$1 & 3 \\ \cline{2-4}
		 & Conv2d & 3$\times$3 & 3 \\ \hline
	\end{tabular}
	\label{Table:convlstm}
\end{table}

Each layer of the \emph{Conv-LSTM} model except the last layer returns the sequences to the next layer and \emph{ReLU} activation has been used in all the layers. The model is trained for 800 epochs with a batch size of 64 by employing Adam optimizer and binary cross-entropy loss.

For non-sequential data the reconstruction is accomplished using only the spatial information present in the occluded frame as there is no temporal information. Therefore, an Autoencoder model is trained for reconstructing the occluded region. The Autoencoder model is trained with binary cross-entropy loss and Adam optimizer for 1000 epochs around which the model is seen to achieve convergence. 


\subsubsection{Fine-tuning the reconstructed frames with \textit{cGAN}}\label{cGAN}
The reconstructed frames by \emph{Conv-LSTM} have minor irregularities in the regions where the actual occlusion was present. 
\textit{cGAN} is used to enhance further by fine-tuning the output of the \emph{Conv-LSTM} model. It takes a single frame as an input generated by the \textit{Conv-LSTM / Autoencoder} and translates similar to ground-truth, thereby generating a fine-tuned or enhanced frame. In this work the output of \textit{Conv-LSTM} is a condition on which the Generator i.e., the default 2D UNet along with the patch-GAN discriminator learns to translate closer to the ground-truth which is the original un-occluded frame. 
We train our \emph{cGAN} for 3000 epochs in 3 sessions by saving checkpoints, with a batch size of 1. The binary cross-entropy loss and L1 loss times $\lambda$ is used to train this model, where $\lambda$ = 100 as shown in \ref{eq2} . 
\begin{equation}
\small
    \min_{G}\max_{D}\mathbb{E}_{x\sim p_{\text{data}}(x)}[\log{D(x)}] +  \mathbb{E}_{z\sim p_{\text{z}}(z)}[1 - \log{D(G(z))}]\\ 
    + \lambda ||L1||
\label{eq2}
\end{equation}
The combination of \textit{ConvLSTM} and \textit{cGAN} produces robust and effective occlusion-free frames.

\subsection{Re-identification using Siamese Network}
Siamese networks 
are particularly effective in predicting whether a given pair of input images are similar or not. 
As in any Siamese network, a pair of input images are passed through two identical channels of convolutional layers, and the difference in the feature embedding is used as a measure of the dissimilarity between the given images. As shown in the figure, in this work, the parallel channels consist of Resnet-101 as basline architecture. Finally, a feature difference block of the two channels is used to get the difference or distance between the embeddings.
Finally, the similarity score between a pair of feature vectors that represent the input images is computed, and hence estimate if the two images are similar or not.

The siamese network is trained with positive and negative pairs of images, where positive pairs are those of same identities, while negative pairs are those of different identities. Adam optimizer with contrastive loss is used to optimize the network parameters. In (\ref{eq6}), $Y$ is the predicted label (either \textit{0} or \textit{1}). Here, \textit{Y=1} indicates the image pairs belong to the same class, whereas \textit{Y=0} indicates that the image pairs are from different classes, $D_{w}$ represents the output score obtained from the softmax layer.
\begin{equation}
\small
    L_{Contr}=(1-Y)\frac{1}{2}(D_{w})^{2}+(Y)\frac{1}{2}\left\{ max(0,m-D_{w}) \right \}^{2}.
\label{eq6}
\end{equation}
The network outputs a score between \textit{0} to \textit{1} depicting whether the given pair of input images belong to the same or different identities. 
We observe that at convergence, the network loss is very low, i.e.,  $10^{-4}$.

\section{Results and Discussion}\label{sec 4}

\begin{figure*}[t]
    \centering
    \includegraphics[width=\textwidth]{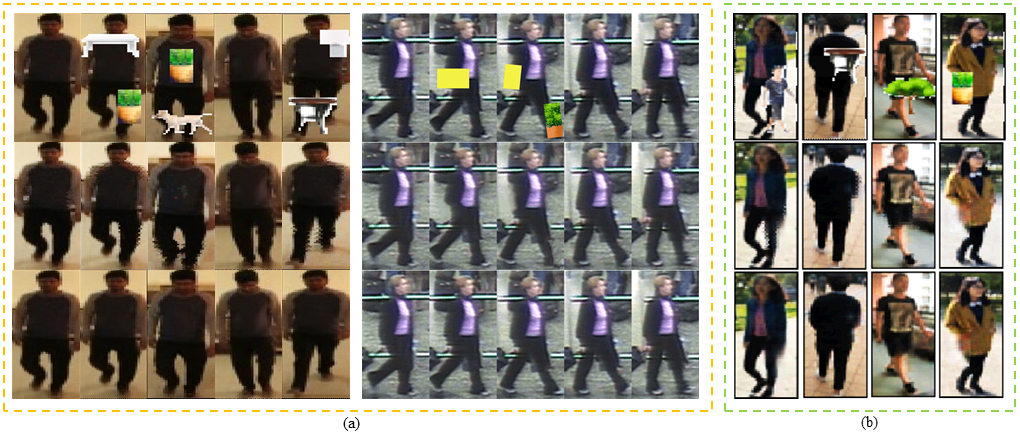} 
    \caption{(a) Occluded sample of sequential frames from a video ($1^{st}$ row), reconstructed frame outputs of \emph{Conv-LSTM} ($2^{nd}$ row), and fine-tuned frame outputs of \emph{cGAN}($3^{rd}$ row), (b) Occluded sample of non-sequential frames i.e, images ($1^{st}$ row), reconstructed frame outputs of \emph{Autoencoder} ($2^{nd}$ row), and fine-tuned frame outputs of \emph{cGAN} ($3^{rd}$ row)}
    \label{res_img}
\end{figure*}

Initially, we start with the dataset description followed by the evaluation settings, experimental results and comparison with existing state-of-the-art approaches.

Four publicly available data sets have been used to test the performance of our proposed approach. The relevant details of each data set, i.e., data set name, number of subjects, if the corresponding dataset is occluded and if it is synthetically occluded in Table \ref{Table:dataset}.
\begin{table}[ht]
\centering
\caption{Description of Datasets used in the study}
\begin{tabular}{|c|c|c|c|}
\hline
\textbf{Data set} & \textbf{No. of} & \textbf{Sequential} & \textbf{Occluded} \\ 
\textbf{Names} & \textbf{subjects} & \textbf{data (Y/N)} & \textbf{data (Y/N)} \\ \hline
\textbf{Occluded ReID} & 200 & N & Y \\ 
\textbf{Partial iLIDS} & 119 & N & Y \\ 
\textbf{iLIDS-VID} & 300 & Y & N \\ 
\textbf{MARS} & 1261 & Y & N \\ \hline
\end{tabular}
\label{Table:dataset}
\vspace{-0.5em}
\end{table}
Among the datasets it is evident that the Occluded ReID and Partial iLIDS are non-sequential (images) while iLIDS-VID and MARS are sequential (videos). For each of the non-sequential data sets mentioned in Table \ref{Table:dataset}, the synthetically occluded frames corresponding to all the subjects are used for training all the models, whereas the images tagged as `occluded' in the dataset are used for testing. 
All the Neural Network models used in the study have been implemented using the Tensorflow framework on a system having 90 GB RAM and one NVIDIA TITAN Xp GPU with two Geforce GTX GPU having 34 GB GPU memory capacity in total.
The evaluation metrics that we use are the mean Average Precision (mAP) and Cumulative Matching Characteristics (CMC).

\begin{table}[ht]
	\centering
	\caption{Effect of Individual Networks in the Reconstruction Framework}
	\begin{tabular}{|c|c|c|c|}
		\hline
		\textbf{Dataset} & \textbf{Network} & \textbf{\emph{Rank-1}}
		 & \textbf{\emph{mAP}}\\ 
		 \hline
		 
		 \textbf{Occluded ReID} & Autoencoder & 79.9 & 68.2 \\ 
		 \textbf{Partial iLIDS} & & 71.3 & - \\ 
		 
		 \hline
		
		 \textbf{iLIDS-VID} & ConvLSTM & 86.4 & - \\ 
		 \textbf{MARS} & & 85.0 & 78.7 \\
		 
		 \hline
		 
		 \textbf{Occluded ReID} & Autoencoder + cGAN & 80.7 & 70.1 \\ 
		 \textbf{Partial iLIDS} & & 73.1 & - \\ 
		 
		 \hline
		 
		 \textbf{iLIDS-VID} & ConvLSTM + cGAN & 87.8 & - \\ 
		 \textbf{MARS} & & 87.5 & 80.6 \\
		 
		 \hline
	\end{tabular}
	\label{Table:resdsc}
\end{table}

In order to evaluate the effectiveness of the individual models in our proposed multi-modal framework we obtain the output frames at each stage and train the siamese network thereby obtaining and comparing the Cumulative Matching Characteristics (CMC) Rank-1 accuracy as shown in Table \ref{Table:resdsc}.

Comparing with the base models in Table \ref{Table:resdsc} it is observed that both the Cumulative Matching Characteristics (CMC) rank-1 accuracy and mean Average Precision (mAP) metrics are significantly influenced by the reconstruction followed by fine-tuning. Our framework improves the Cumulative Matching Characteristics (CMC) rank-1 accuracy of Occluded ReID, Partial iLIDS, iLIDS-VID and MARS datasets from 79.9\% to 80.7\%, 71.3\% to 73.1\%, 86.4\% to 87.8\% and 85.0\% to 87.5\% respectively. While the mean Average Precision (mAP) of both Occluded ReID and MARS improved from 68.2 to 70.1 and 78.7 to 80.3 respectively demonstrating the robustness and effectiveness of occlusion reconstruction.

\begin{table}[ht]
\centering
\caption{Performance comparision of Rank 1 accuracy on Occluded ReID and Partial iLIDS datasets with existing methods}\label{comp1}
\begin{tabular}{|c|c|cc|}
\hline
\textbf{Methods}&\textbf{Occluded}&\textbf{Partial}&\\
&\textbf{ReID}&\textbf{iLIDS}&\\
\hline
     DSR\cite{20}&72.8 (62.8)&58.8&\\
     STNReID\cite{24}&-&54.6&\\
     PVPM\cite{21}&70.4 (61.2)&-&\\
     HOReID\cite{3}&80.3 (70.2)&72.6&\\
     FPR\cite{23}&78.3 (68.0)&68.1&\\
     \hline
     Ours&\textbf{80.7 (70.1)}&\textbf{73.1}&\\
     \hline
\end{tabular}
\end{table}

\begin{table}[ht]
\centering
\caption{Performance comparision of Rank 1 accuracy on MARS and iLIDS-VID datasets with existing methods}\label{comp2}
\begin{tabular}{|c|c|cc|}
\hline
\textbf{Methods}&\textbf{MARS}&\textbf{iLIDS}&\\
&&\textbf{-VID}&\\
\hline
     RRU\cite{26}&84.4 (72.7)&84.3&\\
     STA\cite{25}&86.3 (80.1)&-&\\
     AMEM\cite{28}&86.7 (79.3)&87.2&\\
     A3D\cite{27}&86.3 (80.4)&87.9&\\
     FGRA\cite{29}&87.3 (81.2)&88.0&\\
     \hline
     Ours&\textbf{87.5 (80.6)}&\textbf{87.8}& \\
     \hline
\end{tabular}
\end{table}

A few sample results from our proposed multi-modal reconstruction framework are shown in Fig \ref{res_img}. (a) represents the results of the sequential data while (b) represents the results of non-sequential data. Additionally, the it also includes the frames before and after employing the \emph{cGAN}-based fine-tuning.

Further, we demonstrate a comparative analysis of the proposed re-identification approach with other popular state-of-the-art approaches, namely, \cite{25,20,29,24,26,21,3,23,27,28} in terms of Cumulative Matching Characteristics (CMC) rank-1 re-identification accuracy and mean Average Precision (mAP) represented in brackets '( )' on four public datasets in Table \ref{comp1} and Table \ref{comp2}. Although a few approaches have been developed specifically to work in unoccluded scenarios nevertheless our proposed approach shows superior performance.

\section{Conclusions}\label{sec 6}
In this work, we propose a novel approach for both image (non-sequential) and video-based (sequential) re-identification in the presence of occlusion. A novel occlusion reconstruction framework is proposed which is a combination of \textit{Conv-LSTM} and \textit{cGAN} for video data that uses spatio-temporal information of sequential frames, whereas for image data it is a combination of \textit{Autoencoder} and \emph{cGAN}. After the reconstruction of occluded pixels, a siamese network is used to obtain feature vectors of n-dimensions that are used in computing a similarity score based on which re-identification is executed. Qualitative results show that the approaches proposed for occlusion reconstruction for image and video data are quite effective. Quantitative results by means of re-identification accuracy show that on average our work outperforms the state-of-the-art re-identification approaches, and it performs consistently well for most datasets. This emphasizes the robustness and applicability of our approach in real-time surveillance systems. In future, the generalizability of the proposed framework may be tested on a more extensive open world re-identification databases.

\end{document}